\def\year{2022}\relax
\documentclass[letterpaper]{article} 
\usepackage{aaai22}  
\usepackage{times}  
\usepackage{helvet}  
\usepackage{courier}  
\usepackage[hyphens]{url}  
\usepackage{graphicx} 
\usepackage{natbib}  
\usepackage{caption} 
\usepackage{subcaption}
\usepackage{float}
\DeclareCaptionStyle{ruled}{labelfont=normalfont,labelsep=colon,strut=off} 
\usepackage{algorithm}
\usepackage{algorithmic}

\usepackage{verbatim}

\usepackage[colorinlistoftodos,prependcaption]{todonotes}

\usepackage{newfloat}
\usepackage{listings}
\floatstyle{ruled}
\newfloat{listing}{tb}{lst}{}
\floatname{listing}{Listing}
\title{Fine-grained Intent Classification in the Legal Domain}
\author {
    Ankan Mullick\equalcontrib \textsuperscript{\rm 1}, Abhilash Nandy\equalcontrib \textsuperscript{\rm 1},
    Manav Nitin Kapadnis\equalcontrib \textsuperscript{\rm 2},
    Sohan Patnaik \textsuperscript{\rm 3},
    R Raghav \textsuperscript{\rm 4}
}
\affiliations {
    Indian Institute of Technology Kharagpur, India\\
    \textsuperscript{\rm 1} Department of Computer Science and Engineering \hspace{6mm}
    \textsuperscript{\rm 2} Department of Electrical Engineering\\
    \textsuperscript{\rm 3} Department of Mechanical Engineering \hspace{6mm}
    \textsuperscript{\rm 4} Industrial and Systems Engineering Department\\
    {\tt \{ankanm, nandyabhilash\}@kgpian.iitkgp.ac.in\\
    \{iammanavk,sohanpatnaik106,rraghav5600\}@iitkgp.ac.in}
}

\usepackage{tablefootnote}

\begin{document}

\maketitle
\begin{abstract}
A law practitioner has to go through a lot of long legal case proceedings. To understand the motivation behind the actions of different parties/individuals in a legal case, it is essential that the parts of the document that express an intent corresponding to the case be clearly understood. In this paper, we introduce a dataset of $93$ legal documents, belonging to the case categories of either Murder, Land Dispute, Robbery, or Corruption, where phrases expressing intent same as the category of the document are annotated. Also, we annotate fine-grained intents for each such phrase to enable a deeper understanding of the case for a reader. Finally, we analyze the performance of several transformer-based models in automating the process of extracting intent phrases (both at a coarse and a fine-grained level), and classifying a document into one of the possible $4$ categories, and observe that, our dataset is challenging, especially in the case of fine-grained intent classification.
\end{abstract}

\section{Introduction} 
\label{sec:introduction}



Documents which record legal case proceedings are often perused by many law practitioners. In any Court Judgement, these documents can contain as much as 4500 words (for example - Indian Supreme Court Judgements). Knowing the amount of intent in the text before hand will help a person understand the case better (intent here refers to the intention latent in a piece of text. e.g. ‘Mr. XYZ robbed a bank yesterday’ - in this sentence, the phrase ‘robbed a bank’ depicts the intent of Robbery).

There can be different levels of intent. For example, stating that a legal case deals with murder is a document level intent. It conveys a generalized information about the document. Sentence level and phrase level intents will give much more information about the document. To understand the documents much efficiently various summarization techniques exist. However, an analysis of intents conditioned on the legal cases, along with summarization, would improve the reader's understanding and clarity of the content of the document significantly.


We curate a dataset that consists of 93 legal documents, spread across four intents - Murder, Robbery, Land Dispute and Corruption. We manually annotate certain phrases which bring out the intent of the document. Additionally, we painstakingly assign fine-grained intents (referred to as `sub-intent' interchangeably from here on) to each phrase. These intent phrases are annotated in a coarse (4 categories) as well as in a fine-grained manner (with several sub-intents in each category of intent). For example, under the intent of Robbery, 'Mr. ABC saw Mr. XYZ picking the lock of the neighbour's house' is an example of a witness. Another example is, 'Gold and silver ornaments missing', indicating the stolen items.

Another contribution is the analysis of different off-the-shelf models on intent based task. We finally present a proof-of-concept, which shows that coarse-grained document intent and document classification, as well as fine-grained annotation of phrases in legal documents, can be automated with reasonable accuracy.
\section{Dataset Description} 
\label{sec:dataset}

$5000$ legal documents are scraped from CommonLII~\footnote{\url{http://www.commonlii.org/resources/221.html}} using `selenium' python package. $93$ documents belonging to the categories of Corruption, Murder, Land Dispute, and Robbery are randomly sampled from this larger set. 

Intent phrases are annotated for each document in the following manner - 
\begin{enumerate}
    \item \textbf{Initial filtering:} $2$ annotators filter out sentences that convey an intent matching the category of the document at hand.
    \item \textbf{Intent Phrase annotation} $2$ other annotators then extract a span from each sentence, so as to exclude any details do not contribute to the intent (such as name of the person, date of incident etc.), and only include the words expressing corresponding intent. The resulting spans are the intent phrases. Inter-annotator agreement (Cohen $\kappa$) is 0.79.
    \item \textbf{Sub-intent annotation}: $1$ annotator who is aware of legal terminology, is asked to go through the intent phrases of several documents from all the $4$ intent categories in order to come up with possible set of sub-intents for each intent category, that covers almost all aspects of that category. After coming up with the sets of sub-intents, $4$ annotators are then shown some samples on how to annotate sub-intent for a given phrase. Then, the intent phrases are divided amongst these annotators, and the sub-intent of each intent phrase is annotated thereafter.
\end{enumerate}

Table~\ref{tab:datasets_stats} shows the statistics of our dataset, describing the number of documents, average length of documents and intent phrases, and average sentiment score for each of the $4$ intent categories. The documents on Corruption and Land Dispute are roughly longer than those on Murder and Robbery. Table~\ref{tab:datasets_stats} also shows average sentiment scores across annotated intent phrases (calculated using \emph{sentifish}~\footnote{\url{https://pypi.org/project/sentifish/}} Python Package) for each of the four categories. The sentiment scores of the categories follow the following order - Land Dispute $>$ Corruption $>$ Robbery $>$ Murder, which follows common intuition.

\begin{table*}[t]
\centering
\resizebox{0.8\textwidth}{!}{\begin{tabular}{|l|c|c|c|c|c|}
\hline
\textbf{Category}                                      & \textbf{\begin{tabular}[c]{@{}c@{}}No. of \\ documents\end{tabular}} & \textbf{\begin{tabular}[c]{@{}c@{}}Avg. no. of \\ words/doc\end{tabular}} & \textbf{\begin{tabular}[c]{@{}c@{}}Avg. no. of\\ sentences/doc\end{tabular}} & \textbf{\begin{tabular}[c]{@{}c@{}}Avg. length\\ of intent\\ phrase\end{tabular}} & \textbf{\begin{tabular}[c]{@{}c@{}}Avg.\\ Sentiment Score\\ of intent phrases\end{tabular}} \\
\hline
Corruption                                             & 17                                                                   & 4466                                                                      & 174                                                                          & 17                                                                                & 0.008                                                                                       \\
\begin{tabular}[c]{@{}l@{}}Land Dispute\end{tabular} & 25                                                                   & 4681                                                                      & 186                                                                          & 19                                                                                & 0.02                                                                                        \\
Murder                                                 & 30                                                                   & 2876                                                                      & 135                                                                          & 17                                                                                & -0.012                                                                                      \\
Robbery                                                & 21                                                                   & 2756                                                                      & 118                                                                          & 9                                                                                 & -0.002  \\
\hline
\end{tabular}}
\caption{Statistics for each category in the dataset. The numbers (other than the average sentiment score) are rounded to the nearest integer.}
\label{tab:datasets_stats}
\end{table*}

\begin{figure*}[t]
	\centering
	\subfloat[Corruption]{%
	\centering
		 \includegraphics[width=0.50\textwidth]{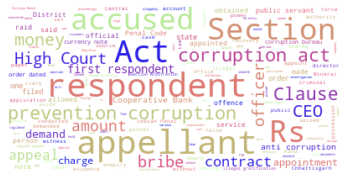} \label{corr_wordcloud}}
	\subfloat[Land Dispute]{%
	\centering
		 \includegraphics[width=0.50\textwidth]{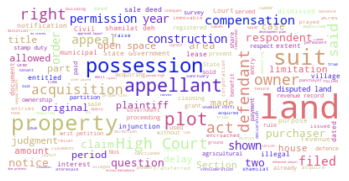} \label{land_wordcloud}}\\
	\subfloat[Murder]{%
	\centering 

		 \includegraphics[width=0.50\textwidth]{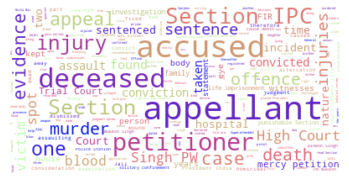} \label{murd_wordcloud}}
	\subfloat[Robbery]{%
	\centering
		 \includegraphics[width=0.50\textwidth]{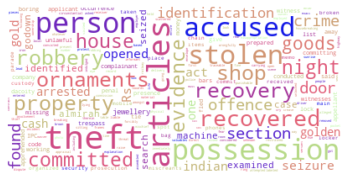} \label{rob_wordcloud}}
	\caption{Wordclouds for each intent category, showing the $200$ most frequently occurring words in the intent phrases for the corresponding category}
	\label{fig:wordcloud}
	\vspace{-2mm}
\end{figure*}

Fig.~\ref{fig:wordcloud} shows the top $200$ most frequent words (excluding stopwords) occurring in the intent phrases for each of the four categories, with the font size of the word being proportional to its frequency. In each wordcloud, we can observe that each category has words that match the corresponding intent (E.g. 'bribe' in Corruption, 'property' in Land Dispute etc.)

\section{Experiment and Results} 
\label{sec:methodology}

This section is organized to describe the use of transformers \cite{vaswani2017attention} for document classification, which will be followed by the explanation for the use of JointBERT \cite{chen2019bert} for intent as well as slot classification. We use two Tesla P100 GPUs with 16 GB RAM to perform all the experiments.

\subsection{Document Classification}

Recent advancements show that, Transformer \cite{vaswani2017attention} based pre-trained language models like BERT \cite{devlin2019bert}, RoBERTa \cite{liu2019roberta}, ALBERT \cite{lan2020albert}, and DeBERTa \cite{he2021deberta}, have proven to be very successful in learning robust context-based representations of lexicons and applying these to achieve state of the art performance on a variety of downstream tasks such as document classification in our case. 

\begin{table}[H]
\centering
\begin{tabular}{|c|c|c|}
\hline
\textbf{Model Name} & \textbf{Accuracy} & \textbf{\begin{tabular}[c]{@{}c@{}}Macro \\ F1-score\end{tabular}} \\
\hline
BERT          & 0.63          & 0.53          \\
RoBERTa       & 0.74          & 0.64          \\
ALBERT        & 0.53          & 0.61          \\
DeBERTa       & \textbf{0.74} & \textbf{0.71} \\
LEGAL-BERT    & \textbf{0.74} & 0.68          \\
LEGAL-RoBERTa & 0.68          & 0.69  \\
\hline
\end{tabular}
\caption{Results of Transformer Models}
\label{tab:baseline_results}
\end{table}

We then implemented different models mentioned in Table \ref{tab:baseline_results}, for learning contextual representations of the documents whose outputs were then fed to a softmax layer to get the final predicted class of the document. Along with this, we also implemented a variant of LEGAL-BERT \cite{chalkidis2020legalbert} and LEGAL-RoBERTa \footnote{https://huggingface.co/saibo/legal-roberta-base} which were pre-trained on large scale datasets of legal domain-specific corpora which in turn led to much better scores than their counterparts pre-trained on general corpora. 

 Recent improvements to the state-of-the-art in contextual language models such as in the case of DeBERTa perform significantly better than BERT. The same is observed from Table \ref{tab:baseline_results} which shows that the Accuracy and Macro F1-score for DeBERTa came to be the highest among the other models, whereas LEGAL-BERT was at par with DeBERTa in terms of Accuracy score. Further, since DeBERTa is trained previously using the disentangled attention mechanism along with an enhanced mask decoder. The training method is same as that of BERT. Owing to the novel attention mechanism used in DeBERTa, it outperforms the other models in terms of both Accuracy and Macro F1-score.
 
LEGAL-BERT on the other hand is pre-trained and further fine-tuned on legal-domain specific corpora, which in turn lead to its state-of-the-art performance on various legal domain specific tasks. In our case, leveraging LEGAL-BERT outperforms other models since the contextual representation is more inclined towards legal matters. 

All of the transformer models were implemented using sliding window attention \cite{sliding_window}, since the document length for all the documents is greater than the transformer maximum token size. They were trained with a sliding window ratio of 20\% over three epochs with learning rate and batch size set at 2e-5 and 32 respectively. The docuemnts in the dataset are randomly split into train, validation and test sets in the ratio of 6:2:2. Note that, when classifying fine-grained intents, we only consider those sub-intents that have atleast $50$ corresponding phrases. 

We report the Accuracy score and Macro average score for each of the model so as to get an intuition on how the state of art transformer-based architectures perform on document classification in the legal domain.

\subsection{JointBERT}

We implemented BERT for joint intent classification and slot filling \cite{chen2019bert} on our dataset. We also replaced the BERT backbone with other transformer-based models such as DistilBERT and ALBERT. Slot Filling is a sequence labelling task, where BIO Tags are for the classes of `Corruption', Land Dispute', `Robbery' and `Murder', and then the intent classification task for those classes. The dataset is prepared in the following manner - 
Since there is a majority of `O' Tags for the slot filling task, only sentences containing an intent phrase, the one before that, and the one after that are used for training to mitigate class imbalance.
Each token has an intent BIO tag and each sentence with an intent phrase has a target intent. We randomly selected 20\% sample for testing, 20\% for validation. Rest 60\% samples were used for training.  

The models were trained over 10 epochs with a batch size of 16, at a learning rate of 2e-5. At each epoch checkpoint, the model was saved and the model with the highest validation accuracy was picked to evaluate on the test set. As can be seen from Table \ref{tab:coarse_grain_results}, BERT proved to be the best model with an Intent Accuracy as well as Intent Macro F1-score of 0.9. 

\begin{table}[h]
\centering
\begin{tabular}{|c|c|c|}
\hline
\textbf{Model Name} & \textbf{\begin{tabular}[c]{@{}c@{}}Intent\\ Accuracy\end{tabular}} & \textbf{\begin{tabular}[c]{@{}c@{}}Intent\\ Macro \\ F1-score\end{tabular}} \\
\hline
BERT       & \textbf{0.90} & \textbf{0.90} \\
DistilBERT & 0.90          & 0.89          \\
ALBERT     & 0.88          & 0.87    \\
\hline
\end{tabular}
\caption{Results on Intent classification}
\label{tab:coarse_grain_results}
\end{table}

Table \ref{tab:intent_classification_best} gives the evaluation metric scores for each intent separately and the analysis provides evidence that the transformer-based models perform poorly on Corruption intent due to the number of ocuments in that category being the lowest, whereas they perform significantly better on other intents.

\begin{table}[H]
\centering
\resizebox{\columnwidth}{!}{\begin{tabular}{|c|c|c|c|c|}
\hline
\textbf{} & \textbf{Precision} & \textbf{Recall} & \textbf{F1-score} & \textbf{Support} \\
\hline
\textbf{Corruption}   & 0.75 & 0.89 & 0.81 & 27  \\
\textbf{Land Dispute} & 0.95 & 0.88 & 0.91 & 42  \\
\textbf{Murder}       & 0.94 & 0.94 & 0.94 & 50  \\
\textbf{Robbery}                              & 0.96 & 0.89                         & 0.92 & 27  \\
\textbf{Macro Average}                        & 0.90 & 0.90                         & 0.90 & 146\\
\hline
\end{tabular}}
\caption{Results of Joint BERT on Intent Classification }
\label{tab:intent_classification_best}
\end{table}

Table \ref{tab:slot_classification_best} enumerates the results of Joint BERT on the task of Slot Classification. The model performs best on Murder intent when compared with others, which is again due to the number of samples in the Murder category being the largest. 

\begin{table}[H]
\centering
\resizebox{\columnwidth}{!}{\begin{tabular}{|c|c|c|c|c|}
\hline
\textbf{} & \textbf{Precision} & \textbf{Recall} & \textbf{F1-score} & \textbf{Support} \\
\hline
\textbf{Corruption}   & 0.74 & 0.38 & 0.51 & 326  \\
\textbf{Land Dispute} & 0.71 & 0.55 & 0.62 & 317  \\
\textbf{Murder}       & 0.80 & 0.63 & 0.70 & 361  \\
\textbf{Robbery}      & 0.66 & 0.53 & 0.59 & 137  \\
\textbf{Macro Average} & 0.73 & 0.52 & 0.60 & 1041\\
\hline
\end{tabular}}
\caption{Results of Joint BERT on Slot Classification }
\label{tab:slot_classification_best}
\end{table}

Table \ref{tab:fine-grained_ic} provides the classification accuracy and Intent Macro F1-score on fine grained Intent Classification task. As the intent becomes more specific, the scores drop significantly, showing that the models are unable to capture the in-depth context of the intent phrases. However, modle with the BERT backbone still performs the best. This can be attributed to the fact, that BERT has the highest number of parameters (~110 million) as compared to ALBERT (~31 million), and DistilBERT (~50 million).

\begin{table}[h]
\centering
\begin{tabular}{|c|c|c|}
\hline
\textbf{Model Name} & \textbf{\begin{tabular}[c]{@{}c@{}}Intent \\
Accuracy\end{tabular}} & \textbf{\begin{tabular}[c]{@{}c@{}}Intent \\ Macro \\ F1-score\end{tabular}} \\
\hline
BERT    & \textbf{0.53}                                                       & \textbf{0.50}                                                                \\
DistilBERT & 0.46                                                                & 0.40                                                                         \\
ALBERT     & 0.48                                                                & 0.47 \\
\hline
\end{tabular}
\caption{Results on fine-grained Intent Classification }
\label{tab:fine-grained_ic}
\end{table}

Table \ref{tab:best_fine-grained_ic} provides the precision, recall and macro F1 Score for fine-grained intent classification for the best performing model among the three models, i.e., JointBERT with a BERT Backbone. The labels are presented in the form of $X\_Y$, where $X$ is an intent (e.g. Robbery), and $Y$ is a fine-grained intent/sub-intent (e.g. action). We observe that, even though the number of training samples per fine-grained class is quite low, performance on the test set is quite good - The F1-Score for all classes is above $0.4$, and except for two classes,  it is above the halfway mark of $0.5$.

\begin{table}[H]
\resizebox{\columnwidth}{!}{\begin{tabular}{|c|c|c|c|c|}
\hline
\textbf{}                           & \textbf{Precision} & \textbf{Recall} & \textbf{F1-score} & \textbf{Support} \\
\hline
\textbf{Corruption\_action}         & 0.46               & 0.60            & 0.52              & 10               \\
\textbf{Land\_Dispute\_action}      & 0.54               & 0.70            & 0.61              & 20               \\
\textbf{Land\_Dispute\_description} & 0.60               & 0.35                                    & 0.44              & 17               \\
\textbf{Murder\_action}             & 0.57               & 0.48            & 0.52              & 25               \\
\textbf{Murder\_description}                                & 0.44               & 0.71                                    & 0.54              & 24               \\
\textbf{Murder\_evidence}                                   & 0.38               & 0.23                                    & 0.29              & 13               \\
\textbf{Robbery\_action}                                    & 0.71               & 0.63                                    & 0.67              & 19               \\
\textbf{Robbery\_description}                               & 0.67               & 0.33                                    & 0.44              & 12               \\
\textbf{Macro Average}                                      & 0.54               & 0.50                                    & 0.50              & 140   \\
\hline
\end{tabular}}
\caption{Results of Joint BERT on fine-grained Intent Classification }
\label{tab:best_fine-grained_ic}
\end{table}

Note that we have not reported the slot classification results for the fine-grained intents. This is because the number of labels becomes almost twice in this case as compared to intent classification (due to the presence of both B and I tags corresponding to each fine-grained intent, and an O class additionally, as we consider BIO tags for annotation). Hence, the number of samples per class is insufficient to learn a good slot classifier.


\section{Discussion}

We observe that, although transformer-based models are performing well in the case of document classification and coarse-grained intent classification, there is a need for better performance in the fine-grained intent classification case. Hence, we argue that our dataset could be a crucial starting point for research on fine-grained intent classification in the legal domain.
\section{Conclusion} 
\label{sec:conclusion}
This paper presents a new dataset for coarse and fine-grained annotation, as well as, shows a proof-of-concept as to how document as well as intent classification can be automated with reasonably good results. We use different transformer-based models for document classification, and observe that DeBERTa performs the best. We try transformer-based models such as BERT, ALBERT and DistilBERT as the backbones of a joint intent and slot classification neural network, and observe that, BERT performs the best among all the three, both in coarse as well as fine-grained intent classification. However, our dataset is challenging, as there is a lot of scope of improvement in the results, especially in fine-grained intent classification. Hence, our dataset could serve as a crucial benchmark for fine-grained intent classification in the legal domain.

\bibliography{aaai22.bib}
\end{document}